\documentclass[journal,twoside,web]{ieeecolor}
\usepackage{generic}
\usepackage{cite}
\usepackage{amsmath,amssymb,amsfonts}
\usepackage{algorithmic}
\usepackage{graphicx}
\usepackage{algorithm,algorithmic}
\usepackage{hyperref}
\usepackage{orcidlink}
\usepackage{multirow}
\hypersetup{hidelinks=true}
\usepackage{textcomp}
\usepackage{hyperref}
\def\BibTeX{{\rm B\kern-.05em{\sc i\kern-.025em b}\kern-.08em
    T\kern-.1667em\lower.7ex\hbox{E}\kern-.125emX}}
\begin{document}

\title{XAI for In-hospital Mortality Prediction\\via Multimodal ICU Data}
\author{Xingqiao Li\orcidlink{0000-0003-3718-8986}, Jindong Gu\orcidlink{0009-0000-0574-0129}, Zhiyong Wang\orcidlink{0000-0002-8043-0312}, Yancheng Yuan\orcidlink{0000-0002-8243-4683}, Bo Du\orcidlink{0000-0002-0059-8458}, and Fengxiang He\orcidlink{0000-0001-5584-2385}
%\thanks{This work is supported in part by the National Natural Science Foundation of China under Grant 62225113. (Corresponding author: B. Du).} 
\thanks{X. Li and B. Du are with %the National Engineering Research Center for Multimedia Software, Institute of Artificial Intelligence, 
School of Computer Science, %and Hubei Key Laboratory of Multimedia and Network Communication Engineering, 
Wuhan University. %, Wuhan, Hubei 430072, China. 
(e-mails: lixingqiao@whu.edu.cn; dubo@whu.edu.cn). }
\thanks{J. Gu is with the Department of Engineering Science, University of Oxford. %, Oxford OX1 3QG, United Kingdom. 
(e-mail: jindong.gu@outlook.com).}
\thanks{Z. Wang is with the School of Computer Science, Faculty of Engineering, the University of Sydney. %, Darlington NSW 2008, Australia. 
(e-mail: zhiyong.wang@sydney.edu.au).}
\thanks{Y. Yuan is with the Department of Applied Mathematics, the Hong Kong Polytechnic University. %, Hung Hom, Hong Kong SAR. 
(e-mail: yancheng.yuan@polyu.edu.hk).}
\thanks{F. He is with the Artificial Intelligence and its Applications Institute, School of Informatics, University of Edinburgh. %, Edinburgh EH8 9AB, United Kingdom. 
(e-mail: F.He@ed.ac.uk).}}
\maketitle

\begin{abstract}
Predicting in-hospital mortality for intensive care unit (ICU) patients is key to final clinical outcomes. AI has shown advantaged accuracy but suffers from the lack of explainability. To address this issue, this paper proposes an {\it eXplainable Multimodal Mortality Predictor} ({\it X-MMP}) approaching an efficient, explainable AI solution for predicting in-hospital mortality via multimodal ICU data. We employ multimodal learning in our framework, which can receive heterogeneous inputs from clinical data and make decisions. Furthermore, we introduce an explainable method, namely {\it Layer-Wise Propagation to Transformer}, as a proper extension of the LRP method to Transformers, producing explanations over multimodal inputs and revealing the salient features attributed to prediction. Moreover, the contribution of each modality to clinical outcomes can be visualized, assisting clinicians in understanding the reasoning behind decision-making. We construct a multimodal dataset based on MIMIC-III and MIMIC-III Waveform Database Matched Subset. Comprehensive experiments on benchmark datasets demonstrate that our proposed framework can achieve reasonable interpretation with competitive prediction accuracy. In particular, our framework can be easily transferred to other clinical tasks, which facilitates the discovery of crucial factors in healthcare research. The code is available at \href{https://github.com/lixingqiao/XAI-ICU}{https://github.com/lixingqiao/XAI-ICU}.
\end{abstract}

\begin{IEEEkeywords}
explainable artificial intelligence, intensive care unit, in-hospital mortality, multimodality
\end{IEEEkeywords}

\section{Introduction}
\label{sec:introduction}
\IEEEPARstart{I}{ntensive} care unit (ICU) patients are highly susceptible to infections and have the highest mortality rate among all hospital units \cite{ref40,ref41}. It is thus crucial to early detect greater risks of death \cite{ref26}. %Due to the explosive growth in the amount of clinical data \cite{ref42}, m
Many deep learning algorithms have been applied to in-hospital mortality prediction and achieved impressive accuracy  \cite{ref27,ref15,ref28,ref17,he2020recent}. However, deep learning algorithms are currently a ``black box'' -- the rationale behind predictions is opaque \cite{ref29}, despite that explaining the decisions is critical to help gain the trust of clinicians and patients \cite{ref34,ref18}.

To improve the explainability of deep learning, researchers have proposed the eXplainable AI (XAI) to provide insights into model behaviour \cite{ref30,ref31}. Whereas much of the work in XAI has been done in natural language processing \cite{ref32} and computer vision tasks \cite{ref33}, XAI for healthcare applications are relatively premature, which distinguish themselves by the dependence of time-series and multimodal data %, such as vital signs from monitors and discrete event sequences from electronic health records (EHRs) 
\cite{ref37}. With the improvement of data capturing in the healthcare system, massive amounts of data are available from different sources, e.g., high-density vital signs from bedside monitors, documented progress notes from caregivers, and discrete event sequences from electronic health records. Integration of these data offers opportunities for developing high performance models and making better clinical decisions \cite {ref14}. However, the heterogeneities across the modalities increase the difficulty of modeling such data. For example, the dense and continuous vital signs can dominate the representations and mask complementary information from notes or discrete event sequences \cite{ref15}. %So, another challenge in this work is to model the different modalities from ICU data jointly. Besides, the interpretation of the multimodal model can highlight the modality relevant to the prediction. 
To the best of our knowledge, few work is seen in the literature giving XAI solutions on clinical, multimodal data. 
Even some works have developed XAI method on single-modality data like clinical notes \cite{ref17} or discrete events sequences \cite{ref13}, the explanation they provide for every modality is independent, but the contribution from different modalities is incompatible.

%On the other hand, different from highlighting features from text or images that can be easily understood, the continuous and categorical features in time-series data can be challenging to interpret by even human experts. %Therefore, designing explainable methods for clinical data is critical to ensure transparency and trust in healthcare applications.

In this article, we propose an explainable, multimodal framework, eXplainable Multimodal Mortality Predictor (X-MMP), which integrates clinical notes, discrete event sequences, and vital signs from a limited time window following ICU patients' admission and predicts the in-hospital mortality. Specifically, X-MMP comprises three transformer-based encoders, which extract the representations from heterogeneous inputs. We combine those representations via a later fusion process and then make decisions upon the fusion. To obtain the explanation of prediction outcomes, X-MMP contains an explainable module, namely Layer-Wise Propagation to Transformer (LRPTrans), which is based on the Gradient $\times$ Input and LRP-rule. We introduce an improved backpropagation rule in the transformer and keep the conservation of attribution in Gradient $\times$ Input. Since all the sub-encoders in our framework are equipped with an explainable module, we can obtain the attribution of any interested input feature. Specifically, the higher attribution indicates the critical contribution of this feature to the specified model's output. Through backpropagation of the gradient, we explicitly account for the attribution of multimodal input features to the model's output. 

To better evaluate our method, we construct a multimodal dataset by extracting the clinical notes, discrete event sequences, and vital signs from the publicly available, benchmark Medical Information Mart for Intensive Care III (MIMIC-III) dataset and the MIMIC-III Waveform Database Matched Subset  \cite{ref59, ref60, ref61}. %, as publicly benchmark datasets for this type of data are currently unavailable. 
We first train our X-MMP on the multimodal datasets as well as multiple models on every single-modal dataset. %and compare the performance of them with several baseline commonly used for time-series and text modeling. 
The experiment results show a comparable accuracy of our transformer-based model to other deep learning methods in mortality prediction. 
We then conducted an ablation study to verify the complement of multiple modalities. 
Further, to verify the effectiveness of our explanation, we conduct perturbation experiments. We perturb the input and track the behaviour of the model when salient features are sequentially removed. Our X-MMP presents an excellent explainable performance in the perturbation study. 
We lastly explore the contribution of multimodal input and find the salient features in each modality. We visualize the most salient features across multimodal input in some dead cases as evidence for the prediction of death. In all the experiments, our framework well utilises the multimodal ICU data and presents competitive performance in mortality prediction. Our method can interpret the prediction outcomes and visualize the features that contribute to the model's decision-making, which significantly helps improve the transparency of mortality prediction and also has the potential to be transferred to other clinical prediction tasks. 
The code is available at \href{https://github.com/lixingqiao/XAI-ICU}{https://github.com/lixingqiao/XAI-ICU}.

\section{Related works}

In this section, we review related works on deep learning in acute care data and the existing explainable approaches for healthcare.

\subsection{Deep learning in acute care data}

With the widespread implementation of digital health record systems in hospitals, there has been an exponential increase in clinical data from in-patients. The increase in available training samples has enabled the application of deep learning algorithms in various clinical applications. Three types of data are typically utilized for acute care prediction: clinical notes from caregivers, discrete event sequences from EHRs, and high-density vital signs from bedside monitors. 

Discrete event sequences encompass time-stamped, nurse-verified physiological measurements (such as hourly documentation of vital signs and medications) and can be easily collected in almost all hospitals. The variables in discrete event sequences can be modeled as either continuous or categorical features. Previous studies have employed Convolutional Neural Networks (CNN) to learn vector representation and predict tasks, including the patient's length of stay (LoS) and in-hospital mortality \cite{ref4, ref5}. Given that the observation values for each patient are recorded sequentially, recurrent networks like Long-Short Term Memory networks (LSTM) and Gated Recurrent Unit (GRU) have been utilized to learn the sequential relationship among observations and predict tasks, including disease diagnosis and patients readmission \cite{ref1,ref2,ref3}. More recently, attention-based models have been proposed for discrete event sequences modeling\cite{ref43}.

Clinical notes provide valuable information about patients' health status and intuitively depict the patients' symptom changes. However, the unstructured, high-dimensional, and sparse nature of text data necessitates effective representation methods for text modeling. Word embedding, a prevalent method for text representation, has been widely used in previous studies. These studies have employed CNN to extract vector representation from clinical notes and classify biomedical article types \cite{ref6,ref7}. Additionally, word embedding methods such as GloVe and Word2Vec have been applied for clinical text modeling \cite{ref44,ref45}. However, these representations only contain local contexts of words, failing to capture the long-range relationship. Recently, Bidirectional Encoder Representations from Transformers (BERT) have been proposed to enhance the performance of numerous NLP tasks \cite{ref46}. In the clinical domain, ClinicalBERT has been introduced to learn word representations from many de-identified notes in health records, surpassing other deep learning methods in de-identification tasks \cite{ref8}. The success of ClinicalBERT has proven the efficiency of the pre-trained language model in clinical text classification. 

High-density vital signs, often sampled every minute or second, are continuously collected from bedside monitors. The vital signs could be humongous, particularly for patients who stay in the ICU for several days. For example, a single vital sign sampled every second can yield 86,000 values in one day. Therefore, a computationally efficient modeling method is required to handle these multi-channel and high-density input signals \cite{ref15}. Vital signs typically include electrocardiogram (ECG), arterial blood pressure (ABP), Respiration rate (RR), fingertip photoplethysmogram (PPG), and other patient-specific signals. Recent studies have applied CNN to classify heart arrhythmia based on short single-lead ECG signals \cite{ref47}. Other research has utilized LSTM with temporal convolutions on the raw vital signs to predict in-hospital mortality \cite{ref49}.

Recent studies \cite{ref10,ref11,ref12} have employed multimodal learning in healthcare, integrating multiple sources of clinical data to improve prediction accuracy. Another study \cite{ref13} proposed a multimodal transformer to integrate discrete event sequences and progress notes for sepsis detection. Other research \cite{ref15} successfully integrated high-density vital signs and discrete event sequences to predict the length of stay and physiologic decompensation during an ICU visit. To the best of our knowledge, few studies have integrated clinical notes, vital signs, and discrete event sequences, partly due to the lack of complete modalities in real datasets. However, it is undeniable that multiple data sources can provide comprehensive information, enabling more robust and accurate predictions for clinical tasks \cite{ref15}.

\subsection{XAI for healthcare}

Despite the significant strides made in the application of deep learning in healthcare, the majority of these algorithms work as black boxes, rendering their predictive logic challenging to decipher. However, a comprehensible interpretation of clinical prediction can enhance the trust of clinicians and patients. Consequently, Explainable AI techniques have been introduced to develop methods that provide insights into the impact of various clinical variables on different prediction tasks.

With the increasing interest in the XAI for healthcare, a previous study  \cite{ref35} reviewed the requirements that XAI must satisfy to provide explanations in the medical field. Another study \cite{ref36} performed an in-depth analysis of three types of explainable methods for medical image models. It is noteworthy that many healthcare applications are predicated on tabular and time-series data. Recent research \cite{ref29,ref37} have provided overviews of XAI methods appropriate for tabular and time-series data in healthcare. Furthermore, Local Interpretable Model-agnostic Explanations (LIME) have been proposed to identify the significant words for predicting the length of stay \cite{ref16}. Some studies \cite{ref17,ref64} have employed shapley additive explanations (SHAP) to identify salient features in discrete clinical events. However, both LIME and SHAP present computational complexity issues and are not suitable for complex neural networks \cite{ref65}. Consequently, saliency methods have been proposed as alternatives. Generally, the approaches for saliency methods can be categorized into gradient-based and attribution-based methods \cite{ref50}. The gradient-based methods use input gradients as an explanation and compute attribution through backpropagation. A previous study used the Integrated Gradient (IG) method to generate explanations over temporal EHRs \cite{ref66}. As for attribution methods, the LRP method begins from the model's output, which is the top attribution. The LRP then propagates attribution from the output backward to the input features \cite{ref23}. In each layer, the incoming attribution must be fully redistributed to the inputs of that layer, and the attribution of input features is used as explanations. Some studies have introduced the LRP algorithm to analyze the attribution of time-series variables to prediction outcomes \cite{ref67,ref68}.

Recently, the transformer has exhibited excellent performance in healthcare domains \cite{ref13,ref19,ref16,ref17}, necessitating the development of interpretive methods to improve their transparency. Some studies \cite{ref13, ref19} have visualized the raw attention map in ClinicalBERT and explored the relationship between input tokens and prediction outcomes. Besides directly using attention maps as explanations, a recent study proposed 'Attention flow' and 'Attention rollout' to combine the attention map from all layers and generate a new attention map as explanations \cite{ref55}. Although the attention map in the transformer can be easily extracted, the influence of salient features in the map on prediction outcomes remains ambiguous \cite{ref20}. Moreover, interpretations based solely on the attention map overlook other modules in transformers, such as layer normalization or skip-connection. A previous study proposed the Gradient*Input (GI) method to explain the predictions made by transformer \cite{ref69}. However, recent research \cite{ref25} found that the gradient in the transformer only reflects the function locally and thus fails to identify the contribution of input features.

\section{Preliminary}

In this section, we introduce the notation and data preprocessing in this study.

\subsection{Notation}

For a cohort of patients, we denote multimodal input as $\{E, C, V\}$. Given an individual ICU visit indexed by $i$, the discrete event sequences $E_i$ can be represented as $E_i = \{e_1,e_2,...,e_L\}\in\mathbb{R}^{L\times D}$, where $L$ is the number of observations after admission and $D$ represent the number of charted clinical event such as hourly measured glucose, mean blood pressure and lab results. The vital signs $V_i$ are continuously monitored waveforms, including but not limited to heart rate and respiratory rate. It has sampled per minute or second and has a similar representation as the time-series of events, $V_i =(v_1,v_2,...,v_M)\in\mathbb{R}^{M\times N}$, where $M$ is the length of vital sign and $M$ represent the number of channels. The unstructured clinical notes $C_i$ are also derived from the electronic health records. It can be represent as $C_i=(c_1,c_2,...,c_r)$, where $c_r$ is the input token derived from tokenizer. In addition, the true label of a patient's in-hospital mortality is $y_i\in \{0,1\}$ (1 indicates died and 0 indicates lived). As for model interpretation, the attribution of each modality can be represented as $\mathcal{R}(T), \mathcal{R}(C), \mathcal{R}(V)$. The attribution of element $x_i$ to outcomes in the network can be expressed as $\mathcal{R}(x_i)$.

\begin{figure*}[t!]
 \centering         
 \includegraphics[width=1\linewidth]{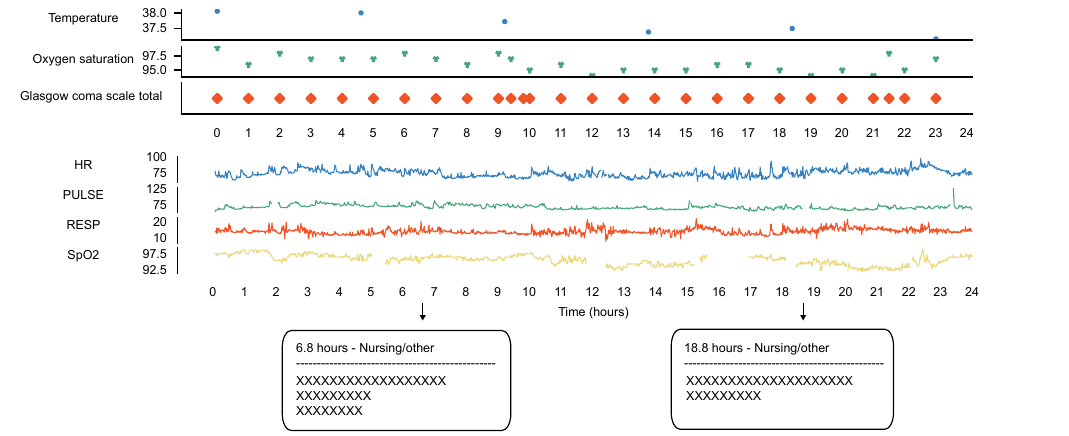}
 \caption{An overview of multimodal ICU data. The discrete event sequences (top) are often recorded irregularly, and they contain categories and continuous features. The vital signs (middle) are multi-channel and high-density. They are often sampled every minute or second. The clinical notes (bottom) are written by the caregiver, and the format and content of notes are different for each patient. The cross is used to substitute the notes for privacy protection.}
 \label{data_demo}
\end{figure*}

\subsection{Data preprocessing}

Our benchmark datasets are derived from the MIMIC-III, a de-identified clinical database comprising 46,520 patients with 58,976 admissions in the ICUs. To extract more information from different modalities, we gathered high-density vital signs from the MIMIC-III Waveform Database Matched Subset. To prepare the in-hospital mortality dataset, we first exclude patients under the age of 18. Additionally, we remove ICU admission records of less than 24 hours or unknown. We select ICU stays without transfer records to prevent multiple admission records for a single patient. The preprocessing steps for each modality are as follows:

\subsubsection{Discrete event sequences} Clinical staff chart vital signs, lab results, and intervention events during hourly rounds and when patients are unstable. Consequently, the observations in the record are often irregular, with the frequency of measurement varying between variables. We limit the observation window to the first 24 hours after patients' ICU admission. The interval of observation time is resampled to once per hour. If multiple values are available for any variables within an hour, we retain the most recently observed values. Missing values are imputed by forward filling. If no previous value is recorded, we use the pre-defined "normal" values for that feature. It is worth noticing that the values in the records are not missing at random. The pattern of missing values can reflect the patient's state. For example, patients with stable conditions will receive less care, resulting in many missing values. Therefore, we model missingness as a feature, as in \cite{ref39}, and provide a binary mask input matrix, indicating whether the time steps contain valid or imputed values. We extract 17 events provided by \cite{ref40} to detect the patient's health state and convert it into a 76-dimensional vector. Specifically, we encode the categorical features in event sequences using one-hot vectors, and we standardize continuous features using zero-mean normalization.

\subsubsection{Clinical notes} For clinical notes, We remove text with error tags and retain the text written in the first 24 hours after patient ICU admission. The types of notes in EHRs include "Nursing" and "Discharge summary." To preserve as many samples as possible, we use all types of notes in this study. We remove special characters and de-identified information in the text. To avoid data leakage, we remove the words related to our prediction targets, such as "die" and "dying." Since our task is to make predictions at the patient level, we concatenate all notes for each patient. We limit the maximum number of words in a patient's note to 512. We start from the end of the 24 hours, working backward until we collect 512 words or exhaust all of the text. This is because the note written later during ICU admission could be more informative than notes written at the beginning \cite{ref56}.

\begin{figure*}[t!]
 \centering         
 \includegraphics[width=1\linewidth]{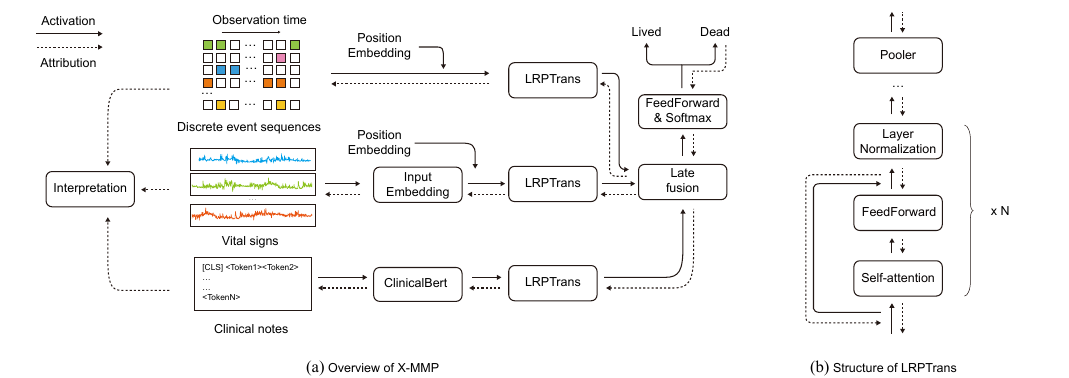}
 \caption{An overview of the X-MMP. The modeling of vital signs and discrete event sequences consists of sequence embedding (input embedding, position embedding), a stack of transformer encoding blocks, and pooler layers. The clinical notes model consists of word embedding from ClinicalBERT and a stack of transformer encoding blocks. The output representations from different modalities are concatenated and fed into a feedforward neural network (FNN). The final softmax layer is used for the prediction. The explanation is based on the backpropagation of the gradient. When the gradient back propagates, the attribution of nodes is calculated by their gradient and value (Gradient $\times$ Input). Finally, we get the attribution of multimodal inputs to the specified prediction. (a) Overview of X-MMP. (b) Structure of LRPTrans.}
 \label{model}
\end{figure*}

\subsubsection{Vital signs} The vital signs from the monitor are multi-channel and high-density. Due to variations in equipment and settings, the vital signs captured from patients can significantly differ. We limit the observation window to the first 24 hours following patients' ICU admission. We extracted 21 types of vital signs from the MIMIC-III Waveform Database Matched Subset. Additionally, if any signs have more than half of their values missing, we discard that sample. To improve the computational efficiency, we resample the frequency to once every three minutes. Thus, a single channel sampled for 24 hours contains 480 numerical values. We impute the missing values with the most recently observed value for the variable. If no previous value is recorded, we also use the pre-defined "normal" values for these signs. All features are standardized using zero-mean normalization. It is worth noting that the bedside monitors are not directly linked to the hospital's digital health system. Therefore, only a subset of the waveform records (34\% in the MIMIC-III Waveforms Database) can be matched with electronic health records. This results in the sample sizes of vital signs being significantly smaller than those of the notes and discrete event sequences.

After data preprocessing, we match the various modalities according to the "ICUSTAY\_ID," which is unique for different ICU admission records. This matched subset contains 4729 multimodal records. The details of multimodal data information are depicted in Fig. \ref{data_demo}.

\section{Method}

In this section, we describe the multimodal modeling for discrete event sequences, clinical notes, and vital signs, followed by a discussion of the explainable approach. 

\subsection{Multimodal modeling}

The structure of X-MMP is illustrated in Fig. \ref{model}.

\subsubsection{Discrete event sequences} For time-series events, the order information is crucial to represent the patient's state. Thus, we apply the sinusoidal functions for positional embedding as in \cite{ref9} to encode the position of the events in the sequence. We then utilize transformer blocks to capture the short- and long-distance dependencies. The multi-head attention can focus on different channels of input features. Following the transformer blocks, we use the pooler layers to extract the first time step of sequences as a representation.

\subsubsection{Clinical notes} For clinical notes, we use the embedding layer of ClinicalBERT to encode the input tokens \cite{ref8}. The ClinicalBERT is a pre-trained language model for medical research. With the adoption of ClinicalBERT, we can represent an input token as a vector, which includes token embeddings and position embedding. After the embedding layers, the matrix of tokens is fed into our transformer encoder. Finally, We use the pooler layers to extract the first [CLS] token as a representation for downstream calculation.

\subsubsection{Vital signs} The vital signs are multi-channel and high-density. Unlike the sole position embedding in event sequences encoding, we add both input embedding and position embedding to encode the raw input. The input embeddings, widely used in NLP tasks, can map low-dimensional vectors to high-dimensional vectors and facilitate sequence modeling \cite{ref58}. For this reason, we apply input embedding in vital signs to capture the dependencies among different channels. We use a linear layer to obtain the high-dimensional embeddings. Additionally, the position embedding is the same as in discrete event sequences. After the embedding layers, we use the transformer blocks to capture long-term dependencies in vital signs. The first time step of the pooler output is extracted as a representation of vital signs.

\subsubsection{Multimodal fusion} The output representations from different modalities are concatenated. The latent representations are fed into a feedforward neural network. We use a softmax layer to predict the probability of in-hospital mortality. In summary, our model is primarily based on the transformer architecture and utilizes the late fusion method to combine the representation of modalities and predict outcomes. 
 
\subsection{XAI method}

In this section, we propose to explain the decision made by our multimodal transformer model. Our explainable method is based on the Layer-wise Relevance Propagation rule (LRP-rule). The principle of LRP-rule posits that the attribution assigned to input variables, which forms the explanation, must sum to the model's output \cite{ref23}. Consequently, the multimodal representation contributes a share of the predicted score at the output:

\begin{equation}\label{eq1}
\mathcal{R}(E_i) + \mathcal{R}(C_i) + \mathcal{R}(V_i)=y_i
\end{equation} 

This principle is known as the conservation axiom. However, transformer heavily rely on skip-connection and attention operators, causing the well-established LRP method to fail in such a model \cite{ref50}. In this study, we incorporate the Gradient $\times$ Input into the LRP-rule, as in \cite{ref25}. Specifically, we replace the well-establish $\beta$-rule with GI for attribution backpropagation. Denote by $e_L$ representing the values from a single observation in discrete event sequences $E_i$, $y_i$ representing probability of death predicted by our multimodal model. The attribution of $e_L$ to $y_i$ can be computed based on the GI method:

\begin{equation}\label{eq2}
\mathcal{R}(e_L)=e_L\cdot(\partial y_i/\partial e_L)
\end{equation}

As previously mentioned, the 'attribution' received by a given component from the above layer will be fully redistributed to the layer below \cite{ref25}. Therefore, the attribution between $\mathcal{R}(e_L)$ and $\mathcal{R}(E_i)$ can be computed as:

\begin{equation}\label{eq3}
\sum_L\mathcal{R}(e_L)=\mathcal{R}(E_i)
\end{equation}

By injecting \eqref{eq1} and \eqref{eq2} into \eqref{eq3}, we obtain the attribution of input features from different modalities to specified prediction outcome:

\begin{equation}\label{eq4}
\sum_L e_L\cdot \frac{\partial y_i}{\partial e_L} + \sum_r c_r\cdot\frac{\partial y_i}{\partial c_r} + \sum_M v_M\cdot \frac{\partial y_i}{\partial v_M}=y_i
\end{equation}

This equation represents the basic combination of LRP and GI on our multimodal transformer network. The attributions of multimodal input features sum to the prediction of a specified class. Furthermore, the attribution of input is calculated by the GI method. This means that we can easily obtain the attribution of input by their gradient and activation values. We can quickly explain the contribution of any interested word in clinical notes or observed values in discrete event sequences through attribution score. 

\subsection{Better LRP rules for transformer}

As mentioned above, the success of LRP-rule is based on the conservation of attribution. The attribution in each layer and node must be fully redistributed to the layer below. However, the self-attention and layer normalization in the transformer can cause a significant break in the conservation and impair the performance of interpretation \cite{ref24,ref25}. To improve the conservation of attribution, we introduce improved gradient propagation rules, as in \cite{ref25}. Specifically, the self-attention and layer normalization are converted to locally linear layers when making an explanation. The implementation of an improved LRP explanation for self-attention can be computed as follows:

\begin{equation}\label{eq6}
A_h = \sum_Le_L[p]_{.detach()}
\end{equation}

The implementation of layer normalization is computed as follows:

\begin{equation}\label{eq7}
L_h = \frac{e_L-\mathbb{E}[e_L]}{[\sqrt{\epsilon+var[e_L]}]_{.detach()}}
\end{equation}

Here, $p$ is the attention map derived from self-attention, and $A_h$, $L_h$ are the output from self-attention and layer normalization, respectively. We use the $detch()$ method in Pytorch to skip the gradient propagation in the attention map and variance in layer normalization. Specifically, this trick creates locally linear expansions of self-attention and layer normalization by treating the attention map and variance as constants. Consequently, these terms can be interpreted as the weights of a linear layer locally mapping the input sequence x to the output sequence y. Through the improved propagation rules, the attribution received from top layers is more conservative, effectively improving our interpretation performance in the transformer.

\section{Experiments}

This section describes the empirical evaluation of our method. The code is available \href{https://github.com/lixingqiao/XAI_medical}{https://github.com/lixingqiao/XAI\_medical}.

\subsection{Experimental implementations}

\subsubsection{Datasets} In this study, we predict the patients' in-hospital mortality based on the data from the first 24-hour ICU stay. We derive the data from the MIMIC-III and MIMIC-III Waveform Database Matched Subset. After the data preprocessing mentioned in the preliminary, we obtained 4,729 multimodal ICU data with 485 in-hospital dead cases (10.2\%). We perform all the experiments using this dataset. The datasets are randomly split, with 80\% for training and 20\% for testing. In addition, we set aside 20\% of the training set for validation. The size of positive and negative samples for all the experiments are shown in Table. \ref{train_test_data_size}.

\begin{table}[htbp]\normalsize
    \centering
    \caption{Summary of training and testing data}
    %\resizebox{\columnwidth}{!}{
    \begin{tabular}{c c c}
		\hline
		& Train & Test\\
        \hline
        Total & 3,547 & 1,182 \\
        Negative & 3,183 & 1,061 \\
        Positive & 364 & 121 \\
		\hline
    \end{tabular}
    \label{train_test_data_size}
\end{table}

\subsubsection{Models \& Optimizer} 

Our model is based on the transformer. The hyperparameter search space for each modality is shown in Table. \ref{hyperparameter}. We perform grid search to optimize hyperparameters. In this research, we formulate in-hospital mortality prediction as a binary classification. Our loss function is based on cross-entropy:

$$Loss(y_i,\hat{y_i})=-y_i\cdot\log(\hat{y_i}) + (1-y_i)\cdot\log(1-\hat{y_i})$$

The training loss is minimized with the Adam optimizer. We also apply learn rate decay with 0.98 per 10 epochs. The area under the receiver operator characteristic curve (AUC-ROC) is a commonly used metric in mortality prediction tasks. In this study, we also report area under the precision-recall curve (AUC-PR) metric since it can be more informative in highly skewed datasets \cite{ref57}. We employ stochastic gradient descent to optimise the loss function \cite{bottou2010large,he2019control}.

\begin{table*}[htbp]\normalsize
    \caption{Hyperparameters search space for each modality}
    \centering
    \label{tab:table3}
    \begin{tabular}{l c c c}
		\hline
		Hyperparameters & Vital signs & Clinical notes  & Discrete event sequences\\
        \hline
        Batch size & [8,16,32] & [8,16,32] & [128,256,512] \\
        Learning rate & [1e$^{-5}$-1e$^{-3}$] & [1e$^{-5}$-1e$^{-4}$] & [1e$^{-5}$-1e$^{-2}$] \\
        Dropout rate & [0.1-0.5] & [0.1-0.5] & [0.1-0.5] \\
        Self-attention layers & [1-4] & [5-10] & [1-6] \\
        Attention heads & [4,8,16,32] & [4,8,16,32] & [4,8,16,32] \\
        Class weights & [1-3] & [1-3] & [1-3] \\
		\hline
    \end{tabular}
    \label{hyperparameter}
\end{table*}

\begin{table*}[t]\normalsize
    \caption{Prediction performance obtained by different models on all modalities}
    \centering
    \setlength\tabcolsep{0.3cm}
    \begin{tabular}{c c c c}
		\hline
		  Modality & Model & AUC-ROC  & AUC-PR\\
        \hline
            \multirow{5}{*}{Discrete event sequences} &  GRU & 0.841 (0.834, 0.848) & 0.425 (0.409, 0.441) \\
            ~ &  LSTM &  0.832 (0.823, 0.842) & 0.396 (0.379, 0.413) \\
            ~ &  RCNN & 0.839 (0.831, 0.848) & 0.426 (0.407, 0.444) \\
            ~ &  AttRNN & 0.836 (0.826, 0.846) & 0.416 (0.394, 0.437) \\
            ~ &  Transformer & 0.842 (0.834, 0.850) & 0.430 (0.411, 0.450) \\
        \hline
            \multirow{5}{*}{Clinical notes} & GRU & 0.558 (0.545, 0.572) & 0.134 (0.126, 0.141) \\
            ~ &  LSTM & 0.580 (0.565, 0.595) & 0.141 (0.131, 0.151) \\
            ~ &  RCNN & 0.763 (0.752, 0.774) & 0.268 (0.255, 0.281) \\
            ~ &  AttRNN & 0.657 (0.635, 0.679) & 0.184 (0.169, 0.200) \\
            ~ &  Transformer & 0.786 (0.776, 0.797) & 0.300 (0.283, 0.316) \\
        \hline
            \multirow{5}{*}{Vital signs} & GRU & 0.676 (0.661, 0.690) & 0.244 (0.233, 0.254) \\
            ~ &  LSTM & 0.678 (0.661, 0.694) & 0.252 (0.240, 0.264) \\
            ~ &  RCNN & 0.725 (0.714, 0.736) & 0.260 (0.243, 0.276) \\
            ~ &  AttRNN & 0.700 (0.688, 0.713) & 0.253 (0.239, 0.267) \\
            ~ &  Transformer & 0.728 (0.717, 0.739) & 0.257 (0.240, 0.274) \\
        \hline
            Vital signs \& Clinical notes & \multirow{3}{*}{Bi-modal} & 0.821 (0.816, 0.826) & 0.351 (0.335, 0.367) \\
            Vital signs \& Discrete event sequences & ~  &  0.849 (0.841, 0.858) & 0.429 (0.404, 0.454) \\
            Clinical notes \& Discrete event sequences &  ~ & 0.851 (0.847, 0.855) & 0.406 (0.390, 0.422) \\
        \hline
            All modalities &  Tri-modal & 0.858 (0.851, 0.866) & 0.430 (0.412, 0.449) \\
        \hline
    \end{tabular}
    \label{performance}
\end{table*}

\subsubsection{Training procedures}

The imbalanced ratio of positive and negative samples is a challenge in in-hospital mortality prediction tasks. So, we up-sample the dead case during training. All the experiments are repeated five times with different random seeds. The five-fold cross-validation (CV) is applied for comparing model performance. We divide the dataset into five equal parts, four of which are used for model training and validation, and the remaining part is used to evaluate the with the same parameters. The training and testing process is repeated five times to ensure that each part of the data participates four times for model training and once for model testing. All experiment results are shown with 95\% confidence interval.

\subsubsection{Hardware and software environments}

All experiments were implemented in PyTorch version 2.0.0, and the code language is Python 3.11.3. Model training is conducted on the Ubuntu Linux release 20.04 system. The CPU is an Intel(R) Xeon(R) CPU E5-2683 v4 @2.10GHz, and the GPU is an NVIDIA Tesla P40. The memory is 22G.

\subsection{Experimental results}

\subsubsection{Prediction performance}

In order to assess the predictive performance, we initially compare single-modal models with four deep learning models that are commonly used for time series and text modeling. These baseline models include GRU, LSTM, attention-based RNN (AttRNN), and recurrent convolutional neural networks (RCNN). Additionally, we conduct a hyperparameter search and select the model with the best performance from all the baselines. Table. \ref{performance} presents the prediction performance of different methods on our benchmark datasets. Our findings reveal that our transformer shows competitive performance over all baselines on AUC-ROC and AUC-PR regardless of modality. In comparison to the baselines, the transformer, owing to the self-attention mechanism, is capable of extracting long-term dependencies from input features. This result validates the efficacy of the transformer in making decisions based on the discrete event sequence, high-density vital signs, and text. It is noteworthy that the input features of discrete event sequences are significantly fewer than the other two modalities. However, the single-modal model predicated solely on discrete event sequences has shown superior performance. This result indicates that the clinical observations are more informative for in-hospital mortality prediction tasks.

\begin{figure*}[htbp]
 \centering
 \includegraphics[width=\linewidth]{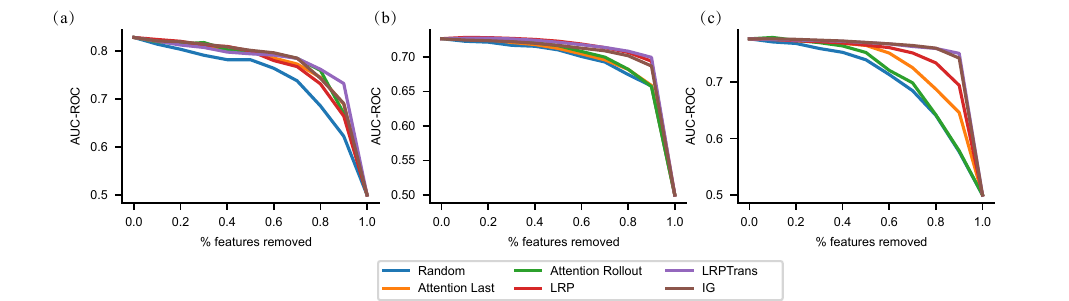}
 \caption{Evaluation of six explainable methods using input perturbations. The input features are sequentially removed based on the absolute attribution from small to large. (a) Discrete event sequences. (b) Vital signs. (c) Clinical notes.}
 \label{perturbation}
\end{figure*}

\begin{table}[t]\normalsize
    \caption{The perturbation study for six explainable methods in different modalities}
    \centering
    \resizebox{\columnwidth}{!}{
    \begin{tabular}{c l c}
		\hline
		  Modality & Method  & AU-AUC-ROC\\
        \hline
            \multirow{6}{*}{Discrete event sequences} &  Random & 0.745 \\
            ~ &  Attention Rollout & 0.772 \\
            ~ &  Attention last &  0.770 \\
            ~ &  IG &  0.774 \\
            ~ &  LRP & 0.767 \\
            ~ &  LRPTrans & 0.777 \\
        \hline
            \multirow{6}{*}{Clinical notes} &  Random & 0.705 \\
            ~ &  Attention Rollout & 0.712 \\
            ~ &  Attention Last &  0.731 \\
            ~ &  IG &  0.755 \\
            ~ &  LRP & 0.744 \\
            ~ &  LRPTrans & 0.755 \\
        \hline
            \multirow{6}{*}{Vital signs} &  Random & 0.693 \\
            ~ &  Attention Rollout & 0.697 \\
            ~ &  Attention Last &  0.696 \\
            ~ &  IG &  0.703 \\
            ~ &  LRP & 0.708 \\
            ~ &  LRPTrans & 0.708 \\
        \hline
    \end{tabular}}
    \label{pertubation-table}
\end{table}

To further explore the impact of multimodal inputs on prediction performance, we perform an ablation study with different combinations of single modality data as inputs. Initially, we consider the performance of models using only bi-modal input. As shown in Table. \ref{performance}, the bi-modal models generally outperform the single-modal model on AUC-ROC. For single-modal models with poor performance, such as those using clinical notes and vital signs, the combination of those two modalities significantly improves the performance. Subsequently, we consider the performance of the complete modality. Our findings reveal that the tri-modal model outperforms all other single- or bi-modal models on both metrics. It is unsurprising that information from additional modalities augments decision-making accuracy. The ablation study suggests that discrete event sequences, vital signs, and clinical notes mutually complement and benefit each other. In the in-hospital mortality prediction task, a model with more available modalities as input performs superiorly than those with fewer modalities.

\subsubsection{Perturbation study}

To verify the performance of our explainable method, we compare it with existing explainable methods using an input perturbation scheme. This scheme involves the removal of input features in ascending order of importance. A robust explainable method should be capable of ranking input features based on their significance, and the elimination of less pertinent nodes should minimally affect the prediction performance. We can evaluate the effectiveness of the explanation by observing the rate of decline in model predictive performance. In this study, our baselines include the raw attention map extracted from the last transformer block, random, attention rollout, and integrated gradients. These methods are commonly used to interpret predictions in transformer-based models. We measure AU-AUC-ROC, which is the area under the AUC-ROC. With the AUC-ROC representing the model's performance after applying the masking to the crucial features, a higher AU-AUC-ROC is desirable as it indicates that removing less relevant nodes has little impact on prediction performance. In table \ref{pertubation-table}, we report the results of the perturbation study in single-modal models. We can see that our proposed improved LRP-based methods have the best interpretation performance across all methods in different modalities. Moreover, the improved LRP method is superior to the basic Gradient $\times$ Input method in discrete event sequences and clinical notes. This result demonstrates the effectiveness of our introduced backpropagation rule. Additionally, explainable methods based on gradient are superior to raw attention-based methods. It is noteworthy that the raw attention-based methods exhibit competitive performance in discrete event sequences and vital signs. Since the number of transformer blocks in the time-series model is fewer than in the text model, other components, such as layer normalization, have little impact on the attribution process. This benefits the explanation method that relies on the raw attention map. In Fig. \ref{perturbation}, we illustrate the perturbation study for different modalities. The perturbation study suggests that the improved LRP explanations are more effective at determining the relevant input features.

\begin{figure*}[htbp]
 \centering
 \includegraphics[width=\linewidth]{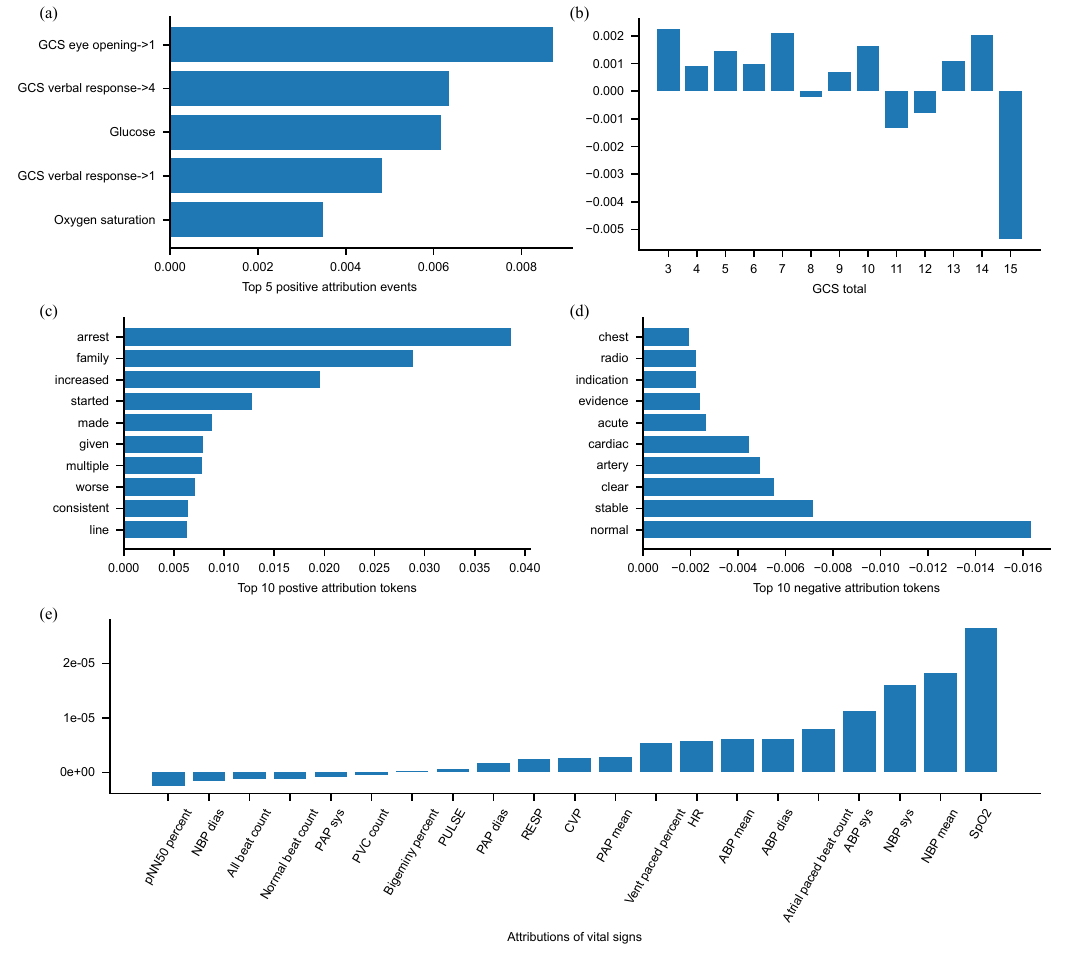}
 \caption{The attribution of input features to the prediction of death in different modalities. A total of 485 in-hospital dead cases are used for analysis. (a) Top 5 positive attribution events in discrete event sequences. (b) Attribution of all "GCS total" scores. (c) Top 10 positive attribution tokens. (d) Top 10 negative attribution tokens. (e). Attribution of all vital signs.}
 \label{salient_features}
\end{figure*}

\subsection{Explanations of predictions}

\begin{figure*}[htbp]
 \centering
 \includegraphics[width=\linewidth]{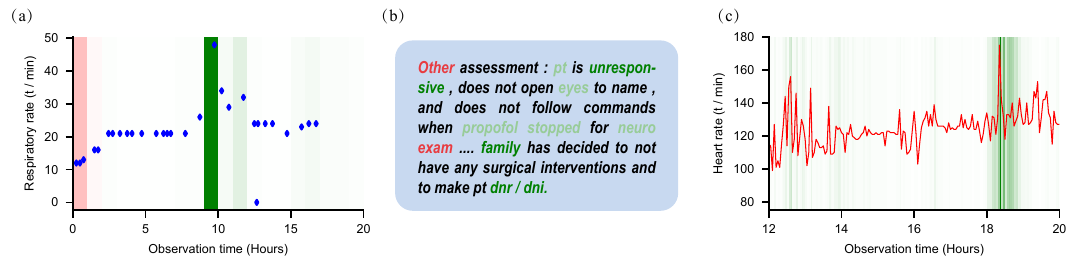}
 \caption{The visualization of the most salient features in three different cases. The darker green color indicates that the features contribute positively to the prediction of death, and red indicates that the feature has a negative contribution. (a) Case with the most salient feature in discrete event sequences. (b) Case with the most salient feature in clinical notes. (c) Case with the most salient feature in vital signs.}
 \label{case_study}
\end{figure*}

\subsubsection{Salient features in in-hospital mortality prediction}

To further identify key factors influencing in-hospital mortality prediction, we applied our explainable method to the tri-modal model. A total of 485 dead cases in the test dataset are used to calculate the attribution of multimodal input features to the prediction of death. A feature with higher attribution indicates its significant impact on prediction. The attribution of features for each modality is depicted in Fig. \ref{salient_features}, with variables arranged according to their mean values. For discrete event sequences, we highlight the top five input features contributing to in-hospital mortality. These features involve in critical clinical information, such as mental state, as measured by the Glasgow Coma Scale (GCS). Notably, "The GCS eye opening-1" implies that the patients have no response to stimulation, indicating a potentially critical condition. Furthermore, we explore the correlation between the "GCS total" score and its attribution to death. We found that a perfect score (15) negatively contributes to death, while a "GCS total" between 3 and 7 consistently contributes positively to death. This result is consistent with medical knowledge that a lower "GCS total" indicates a more severe patient state. In terms of clinical notes, we display the top ten positive and negative attributions tokens. It is worth noticing that due to the ClinicalBERT tokenizer, we eliminated the tokens containing "\#" and restricted the number of occurrences to more than 100 across all patient texts. We find that "arrest" is the most salient word contributing to death. In fact, "arrest" is closely associated with "cardiac arrest," indicating the severity of the patient's condition. Conversely, words such as "normal," "stable," and "clear" negatively contribute to death. For high-density vital signs, we also noted that the "SpO2" significantly contributes to death. However, half of the input features in vital signs showed little attribution to the prediction. This could account for the poor performance of predictions based on vital signs.

\subsubsection{Case studies}

In this section, we perform case studies to qualitatively assess our explanations and visualize the salient features in mortality prediction. We select three dead cases, each displaying the most notable features in different modalities. The results are depicted in Fig. ???\ref{case_study}. In the first case, the most salient feature is the respiratory rate, which is found in discrete event sequences. After 10 hours of ICU admission, we observed that the respiratory rate rapidly increased from 20/min to 50/min. An abnormally high respiratory rate indicates the patient's unstable condition, particularly in an ICU setting. Our explainable methods also highlight the respiratory rate of 50/min. In the second case, the salient features are found in clinical notes. Our explanation method highlights the terms "dnr/dni" and "unresponsive". The "unresponsive" indicates the patient's deteriorated mental state and their lack of response to environmental stimuli. Additionally, "dnr/dni" means that the patient or their family has declined further interventions. Both terms are highly correlated with mortality. In the final case, the significant features are identified as heart rate in vital signs. Bedside monitors recorded abnormal peaks approximately 18.3 hours after the patient's ICU admission, with a heart rate nearing 160/min, which is beyond the normal range. This value may indicate a deterioration in the patient's condition. Our explanatory method also accurately highlights this abnormal peak.

Through visualization, we analyze the significant features in several patients and understand the reasoning behind decision-making. The input features that our explainable method emphasizes are equally deserving of attention in medical research. These results demonstrate the excellent performance of our explainable method in individual cases.

\section{Conclusion}

This paper proposes an { eXplainable Multimodal Mortality Predictor} ({ X-MMP}) approaching an efficient, explainable AI solution for predicting in-hospital mortality via multimodal ICU data. A multimodal transformer is designed to process heterogeneous, clinical data and make decisions. An explainable method, { Layer-Wise Propagation to Transformer}, is designed based on the LRP method, producing explanations over multimodal inputs and revealing the salient features attributed to prediction. The contribution of each modality to clinical outcomes can be visualized, assisting clinicians in understanding the reasoning behind decision-making. Extensive experiments are constructed on a multimodal dataset constructed based on the MIMIC-III dataset and the MIMIC-III Waveform Database Matched Subset. The empirical results demonstrate that our X-MMP can achieve reasonable interpretation with competitive prediction accuracy. Our framework can also be easily transferred to other clinical tasks, facilitating the discovery of crucial factors in healthcare research. The code is available at \href{https://github.com/lixingqiao/XAI-ICU}{https://github.com/lixingqiao/XAI-ICU}.

\section*{References}

\bibliographystyle{IEEEtran}

\bibliography{ms.bbl}

\end{document}